\documentclass[conference]{IEEEtran}
\IEEEoverridecommandlockouts

\usepackage{cite}
\usepackage{amsmath,amssymb,amsfonts}
\usepackage{algorithmic}
\usepackage{graphicx}
\usepackage{textcomp}
\usepackage{xcolor}
\usepackage{multirow}
\usepackage{pifont}
\usepackage{booktabs}
\usepackage{balance}
\usepackage{array}
\usepackage{url}
\usepackage[hidelinks]{hyperref}
\def\BibTeX{{\rm B\kern-.05em{\sc i\kern-.025em b}\kern-.08em
    T\kern-.1667em\lower.7ex\hbox{E}\kern-.125emX}}

\begin{document}

\title{PARAGraph: Pathology-Anatomy-Aware Hierarchical Graph for Diabetic Retinopathy Grading}

\author{
\IEEEauthorblockN{
Ziyang Zhang\IEEEauthorrefmark{1},
Yuankai Huo\IEEEauthorrefmark{1},
Yalin Zheng\IEEEauthorrefmark{2},
He Zhao\IEEEauthorrefmark{2}
}

\IEEEauthorblockA{
\IEEEauthorrefmark{1}
\textit{Vanderbilt University}, USA
}

\IEEEauthorblockA{
\IEEEauthorrefmark{2}
\textit{University of Liverpool}, UK
}

}

\maketitle
\begin{abstract}
Diabetic retinopathy (DR) remains a leading cause of vision loss among working-age adults worldwide, making reliable severity grading clinically important. Despite strong performance, most deep models formulate DR grading as image-level classification and do not explicitly model clinically grounded evidence, such as lesion types and spatial relations. 
% Lesion-based graph methods provide more interpretable reasoning, but they are sensitive to imperfect segmentation.
In this paper, we propose PARAGraph, a Pathology-Anatomy-Aware Hierarchical Graph framework for DR grading. 
PARAGraph represents each image as a three-level hierarchical graph with lesion-level nodes, intermediate category and region nodes, and global anatomical and semantic nodes.
% including lesions, per-class categorical aggregations, spatial regional aggregations, an optic-disc anatomical node, and image-level semantic node. 
To incorporate medical priors into nodes, we construct an optic disc–fovea-anchored coordinate frame that provides a scale- and rotation-normalized retinal reference system. 
Within this frame, lesion nodes are encoded with category, normalized area, and anatomical coordinates. 
% Furthermore, this coordinate frame is used to derive anatomical positional embeddings that are injected into the visual backbone during feature extraction, enabling anatomy-aware feature extraction for subsequent lesion-level graph reasoning.
% Lesion nodes are enriched with explicit medical priors, including lesion type, area, and position in an optic-disc-anchored coordinate frame. This anatomical frame is also injected into the visual backbone as a scale- and rotation-invariant positional encoding. To complement lesion-grounded reasoning, PARAGraph further fuses whole-image context at two levels: early fusion through the semantic node during message passing, and late fusion through an independent global prediction branch. 
To mitigate noisy lesion segmentation, PARAGraph uses a dual-fusion strategy that introduces global visual context into a graph semantic node and a decision-level prediction branch, improving robustness when lesion evidence is unreliable.
Extensive experiments on Messidor-2, APTOS, and DDR show that PARAGraph achieves consistent DR grading performance over state-of-the-art methods. 
Interpretability and robustness analyses further demonstrate that its predictions are clinically grounded, closely associated with lesion evidence and robust to lesion segmentation noise. The code will be released at \url{https://github.com/Ken-zzzzy/PARAGraph}.
\end{abstract}

\section{Introduction}

Diabetic retinopathy (DR) is a microvascular complication of diabetes. It is a leading cause of vision impairment among working-age adults worldwide~\cite{diabetic}. Approximately 34.6\% of people with diabetes develop DR~\cite{global}. This highlights the need for reliable severity grading to support risk stratification and treatment planning.

Deep learning has driven substantial progress in automated DR grading. However, two major paradigms still leave important gaps. Image-level classifiers~\cite{cabnet,development,automated} achieve strong accuracy, but they often operate as black boxes. Lesion evidence is encoded only \emph{implicitly} within global features. As a result, predictions cannot be easily traced to identifiable pathology. This also makes the model design less aligned with the lesion-grounded reasoning used by clinicians.
Structure-aware methods instead model class dependencies~\cite{green} or build graphs over detected lesions~\cite{graphbased}. Lesion-based graphs provide explicit and inspectable evidence, but their reliability depends heavily on the quality of their inputs. Lesion detection remains difficult in real-world fundus images~\cite{idrid}. When a graph is constructed solely from detected lesions, missed lesions lead to incomplete graph structures, while false detections introduce spurious nodes. These detection errors introduce structural noise and can compromise stability and generalization.

We argue that a trustworthy grader should make lesion evidence \emph{explicit} while remaining \emph{robust} to imperfect detection. To this end, we propose \textbf{PARAGraph} (\textbf{P}athology–\textbf{A}natomy-Aware Hie\textbf{ra}rchical Graph), a within-image heterogeneous graph for DR grading. PARAGraph represents clinically meaningful evidence through a hierarchy of lesion, intermediate and global nodes. 
% It further integrates explicit medical priors into node features.
% It also incorporates anatomical guidance and whole-image context to improve robustness. 
To make lesion nodes clinically informative, we construct an optic disc–fovea-anchored coordinate frame, from which lesion location and scale-normalized area are derived and combined with lesion-category information.
Beyond lesion-level priors, PARAGraph further integrates whole-image context and confidence-weighted message passing to reduce the impact of imperfect segmentation while preserving lesion-grounded transparency.
% An optic-disc–anchored positional encoding provides a consistent anatomical reference frame for visual features. This helps the model reason about lesion evidence in relation to retinal anatomy. In addition, global image information is fused with graph reasoning, providing a holistic fallback when lesion evidence is sparse or noisy. Confidence-weighted edges further reduce the influence of unreliable detections. In this way, PARAGraph retains the transparency of lesion-based reasoning while reducing its fragility.
Our main contributions are:
\begin{itemize}
\item We propose \textbf{PARAGraph}, a pathology-anatomy-aware hierarchical graph for DR grading that integrates clinical priors into lesion- and region-level reasoning.
% \item We design an \textbf{optic disc–anchored positional encoding} that encodes each location by its polar coordinates relative to the disc, injecting a scale and rotation invariant anatomical reference frame into the visual backbone.
\item We design an \textbf{optic disc–anchored positional anatomical coordinate frame} that provides scale and rotation invariant locations for deriving lesion-level priors and anatomy-aware positional embeddings.
\item We introduce a \textbf{two-stage global fusion} that injects global context into the graph at different stages, complementing lesion-level evidence with holistic context.
\end{itemize}

\section{Related Work}

\subsection{End-to-End DR Grading}

Most deep DR grading methods map fundus images directly to severity labels. Early approaches adopted CNN backbones such as DenseNet~\cite{densenet} and EfficientNet~\cite{efficientnet}, followed by attention-based architectures~\cite{cabnet,focused}, vision transformers~\cite{vit}, and self-supervised foundation models~\cite{retfound}. Post hoc methods such as CAM and Grad-CAM~\cite{cam,gradcam} provide visual explanations of model predictions. However, these methods still encode lesion evidence implicitly within image features, without explicitly modeling clinically relevant attributes such as lesion type, size, count, and spatial distribution. This limits their interpretability and alignment with clinical reasoning.

\subsection{Structure- and Lesion-Aware Grading}

Graph-based approaches seek to model pathological relationships explicitly. Class-dependency graphs~\cite{green} lack direct pathological correspondence, while image-level similarity graphs~\cite{qagcn} capture inter-image rather than intra-retinal relationships. Other clinically grounded approaches employ neuro-symbolic reasoning~\cite{xai} or construct graphs from lesion segmentations~\cite{graphbased}. PRANet~\cite{pranet} models pathological co-occurrence through a pathological relationship perception module and dual attention, while MDGNet~\cite{mdgnet} combines multi-scale dynamic feature fusion with visual graph convolution.

However, graph-based methods relying on explicit lesion detection or segmentation face two key limitations. First, lesion-only graphs are restricted to predefined detectable categories, potentially overlooking diffuse abnormalities, vascular changes, and broader contextual evidence. Second, their performance depends heavily on upstream lesion detection, which remains challenging for small, low-contrast, and imbalanced lesions~\cite{idrid}. Missed or spurious detections alter graph topology, introducing incomplete or unreliable evidence that conventional pipelines cannot effectively mitigate.

To address these limitations, PARAGraph introduces an anatomy-aware heterogeneous graph that integrates explicit lesion evidence, medical priors, and global image context. Optic-disc–anchored positional encoding establishes a consistent anatomical reference frame, while contextual integration reduces dependence on imperfect lesion segmentation. This design aims to preserve clinically grounded interpretability while improving grading robustness.

\section{Method}
Given a dataset $\mathcal{D}= { (x_i,y_i)}_{i=1}^{N}$, where $x_i \in \mathbb{R}^{W \times H \times 3}$ is the color fundus image, $y_i \in \{0,1,2,3,4 \}$ denotes the DR grade label. Our goal is to learn a mapping function that predicts DR severity for unseen test samples $\mathcal{D}_t$. To this end, for each image we construct a hierarchical graph $\mathcal{G}= \{ \mathcal{V}, \mathcal{E} \}$, where nodes $\mathcal{V}$ encode lesion/intermediate/global-level representations, and edges $\mathcal{E}$ model the interactions. 

\begin{figure*}[!t]
\centering
\includegraphics[width=0.9\linewidth]{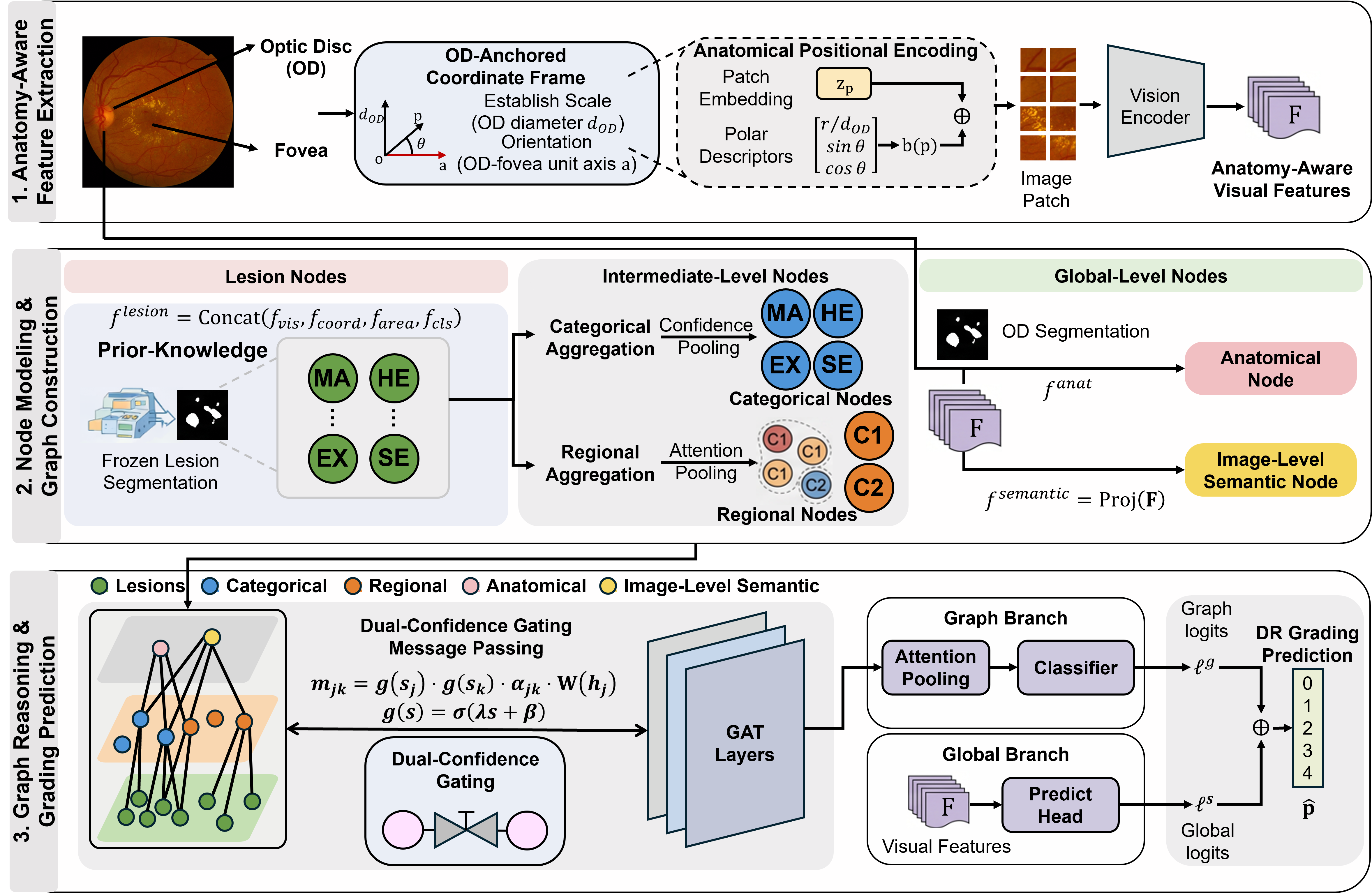}
% \caption{Overview of the proposed PARAGraph framework. Lesion instances extracted from a frozen segmentation backbone are enriched with anatomy-aware priors and organized into a hierarchical graph. Global context fusion and confidence-gated message passing further complement lesion-grounded reasoning for robust DR grading.}
\caption{
Overview of the proposed PARAGraph framework.
(1) An OD-anchored anatomical coordinate frame defines a unified anatomical reference system and produces anatomy-aware positional descriptors for visual feature extraction.
(2) Lesion instances from a frozen segmentation backbone are enriched with medical priors and aggregated into lesion, categorical, and regional nodes, while OD segmentation and the image-level representation form the anatomical and semantic nodes, respectively.
(3) Confidence-gated graph reasoning and a global branch combine lesion-grounded evidence with holistic image context for robust DR grading.
}
\label{fig:framework}
\end{figure*}

PARAGraph is built on three key designs: 
% (i) an \emph{OD-anchored anatomical positional encoding} that injects a scale- and rotation-invariant reference frame into the visual encoder (Sec.\ref{sec:ope});
(i) an \emph{OD-anchored anatomical coordinate frame} that provides scale and rotation invariant locations for lesion-aware modeling and feature extraction (Sec.\ref{sec:ope});
(ii) a \emph{prior-knowledge-enhanced hierarchical graph} that organizes lesion evidence into clinically grounded multi-level nodes (Sec.\ref{sec:graph}); and (iii) an \emph{early–late global fusion} that integrates whole-image context with lesion-grounded reasoning (Sec.\ref{sec:fusion}). The graph is further regulated by a dual-confidence gating mechanism, and a Graph Attention Network (GAT)\cite{gat} performs structured message passing for DR grading (Fig.\ref{fig:framework}).

% \subsection{OD-Anchored Anatomical Positional Encoding}
\subsection{OD-Anchored Anatomical Coordinate System }
\label{sec:ope}

The optic disc (OD) and fovea are clinically meaningful anatomical landmarks in fundus images. Although different images may contain different lesion patterns, these two anatomical structures are consistently present across fundus images. They therefore provide a natural basis for defining a unified anatomical reference frame for DR grading. We build an anatomy-aware coordinate system anchored by the OD and fovea.

This coordinate system provides normalized anatomical coordinates for two purposes: deriving lesion-level priors, including location and scale-normalized area, and generating positional encoding for anatomy-aware feature extraction in the visual backbone.
Specifically, the OD center is defined as the origin, and the direction from the OD center to the fovea center defines the positive $x$-axis. To improve robustness across datasets and imaging scales, spatial coordinates are normalized by the OD diameter.
Each lesion is encoded by its anatomical location under the coordinate frame, together with its scale-normalized area.
Beyond lesion-node prior construction, the same coordinate system is also used to define patch-wise anatomical descriptors for positional embedding in the visual backbone.

Formally, let $\mathbf{o}$ and $\mathbf{c}_{f}$ denote the OD and fovea centers, and let $d_{OD}$ denote the OD diameter. For a patch located at $\mathbf{p}$, we define the displacement from the OD center as $\boldsymbol{\Delta}=\mathbf{p}-\mathbf{o}$. Its radial distance is $r=\lVert\boldsymbol{\Delta}\rVert_{2}$. We also define the OD--fovea unit axis as
\begin{equation}
\mathbf{a}=\frac{\mathbf{c}_{f}-\mathbf{o}}{\lVert\mathbf{c}_{f}-\mathbf{o}\rVert}.
\end{equation}
The orientation of $\mathbf{p}$ relative to this anatomical axis is represented by
\begin{equation}
\cos\theta=\frac{\boldsymbol{\Delta}\cdot\mathbf{a}}{r},
\qquad
\sin\theta=\frac{\Delta_{x}a_{y}-\Delta_{y}a_{x}}{r},
\end{equation}
where the signed sine term distinguishes the two sides of the OD--fovea axis.

We then form a scale- and rotation-invariant polar descriptor:
\begin{equation}
\mathbf{e}(\mathbf{p})=
\Big[
\tfrac{r}{d_{OD}},\
\sin\theta,\
\cos\theta
\Big].
\end{equation}
This descriptor is mapped through an MLP to obtain an anatomical positional bias:
\begin{equation}
\mathbf{b}(\mathbf{p})=\mathrm{MLP}\big(\mathbf{e}(\mathbf{p})\big).
\end{equation}
The bias is injected into the visual backbone by adding it to the corresponding patch embedding:
\begin{equation}
\tilde{\mathbf{z}}_{\mathbf{p}}=\mathbf{z}_{\mathbf{p}}+\mathbf{b}(\mathbf{p}).
\end{equation}

It allows the encoder to represent image regions in an OD-anchored coordinate system. This provides more consistent spatial cues across datasets and imaging conditions. All subsequent node visual features, including lesion, anatomical, and global features, are extracted from this anatomy-aware encoder.

\subsection{Prior-Knowledge-Enhanced Hierarchical Graph}
\label{sec:graph}

The proposed graph $\mathcal{G}$ comprises three levels of nodes $\mathcal{V}$: lesion-level, intermediate-level, and global-level nodes. The lesion, intermediate, and anatomical nodes are detailed below, while the image-level semantic node is introduced in Sec.~\ref{sec:fusion}.

\subsubsection{Lesion Node Feature Modeling}
Each detected lesion is represented as a node combining visual appearance with clinically grounded priors. Lesions are localized by the frozen segmentation model, and their visual embedding $f_{vis}$ is sampled from the anatomy-aware encoder features (Sec.~\ref{sec:ope}). The normalized lesion location is encoded as
\begin{equation}
\label{eq:emb_coor}
f_{coord} = \phi(u_j,v_j),
\end{equation}
where $(u_j,v_j)$ denote the normalized coordinates of lesion $j$, and $\phi(\cdot)$ projects them into the embedding space.

Since lesion extent is clinically informative~\cite{diabetic}, we encode its area as
\begin{equation}
\label{eq:emb_area}
f_{area} = \rho(a_j), \qquad a_j = \frac{\mathrm{Area}_{j}}{d_{OD}^{2}},
\end{equation}
where $\mathrm{Area}_{j}$ denotes the lesion pixel area, normalized by the squared OD diameter $d_{OD}^{2}$ for scale consistency (Sec.~\ref{sec:ope}), and $\rho(\cdot)$ is the projection function.

To preserve lesion-type information, we define
\begin{equation}
\label{eq:emb_cls}
f_{cls} = \eta(y_j),
\end{equation}
where $y_j$ and $\eta(\cdot)$ denote the lesion category and projection function, respectively. These embeddings are concatenated with $f_{vis}$ to form the prior-enhanced lesion representation:
\begin{equation}
\label{eq:emb_node}
f^{lesion} = \mathrm{Concat}(f_{vis}, f_{coord}, f_{area}, f_{cls}).
\end{equation}

\subsubsection{Intermediate-Level Representation Aggregation}
We introduce categorical and regional nodes to aggregate lesion information from semantic and spatial perspectives.

\textit{\textbf{Categorical nodes}} aggregate lesions of the same category through confidence-weighted pooling of their visual embeddings, emphasizing reliable lesion evidence without introducing additional learnable aggregation parameters. Their representations are subsequently refined through graph-attention message passing with connected lesion nodes.

\textit{\textbf{Regional nodes}} capture spatial lesion clustering in the anatomy-aware coordinate space. Lesions are grouped by K-means~\cite{kmeans} using coordinates $p_j$, and each cluster is aggregated via multi-head attention pooling~\cite{attention},
where a learned region token queries the lesion features. This enables cross-category interaction among spatially adjacent lesions.

Together, these intermediate nodes capture category-level semantics and regional spatial patterns for hierarchical graph reasoning.

\subsubsection{Anatomical and Image-Level Semantic Modeling}
Since lesion-based representations depend on segmentation quality, we introduce two global-level nodes to provide complementary context and improve robustness.

\textit{\textbf{Anatomical node}} serves as the graph's anatomical anchor. The optic disc (OD) defines the origin of the anatomy-aware coordinate system, while its diameter provides scale normalization across images. This establishes a stable reference frame independent of lesion segmentation. Multi-scale visual features extracted from the OD region are aggregated into the anatomical representation $f^{anat}$, providing structural context for graph reasoning.

\textit{\textbf{Image-level semantic node}} captures holistic visual information beyond segmented lesions, complementing the anatomical anchor. Its representation is detailed in Sec.~\ref{sec:fusion}.

\subsection{Global Context Fusion}
\label{sec:fusion}
Lesion and intermediate nodes encode pathology through segmented lesions. Their reasoning is therefore bounded by the quality and completeness of lesion segmentation. However, fundus images may contain rich contextual cues and disease-related abnormalities that are diffuse, difficult to annotate, or absent from the available segmentation categories.

To better exploit full-image information and maintain reliable grading when segmentation is incomplete or inaccurate, we introduce global context at two complementary stages. Early fusion injects image-level semantics into the graph as a global semantic node, allowing whole-image cues to participate in message passing and shape lesion-grounded representations. Late fusion uses an independent branch at the decision level, preserving a holistic prediction path that does not rely on the lesion graph topology.

\subsubsection{Semantic Node and Early Fusion}

Let $\mathbf{F}$ denote the global representation produced by the anatomy-aware encoder over the whole image. The semantic node feature is obtained by a projection:
\begin{equation}
\label{eq}
f^{semantic} = \mathrm{MLP}(\mathbf{F}),
\end{equation}
which captures high-level semantic abstraction beyond explicitly segmented lesions.
As a graph node, the semantic node participates in message passing with lesion, intermediate, and anatomical nodes. It introduces complementary image-level cues into the lesion-grounded reasoning process, thereby enriching the graph representation before prediction.

\subsubsection{Late Fusion}

Beyond serving as a graph node, the global representation $\mathbf{F}$ also drives an independent global branch:
\begin{equation}
\label{eq:logit}
\boldsymbol{\ell}^{s} = \mathrm{Head}_{s}(\mathbf{F}),
\end{equation}
where $\boldsymbol{\ell}^{s}$ denotes the global-branch logits. The graph branch produces logits $\boldsymbol{\ell}^{g}$ through the gated graph reasoning described in Sec.~\ref{sec:reason}. We combine the two branches at the decision level:
\begin{equation}
\label{eq:fusion}
\boldsymbol{\ell} = \boldsymbol{\ell}^{g} + \boldsymbol{\ell}^{s},
\qquad
\hat{\mathbf{p}} = \mathrm{softmax}(\boldsymbol{\ell}).
\end{equation}
This late fusion preserves the lesion-grounded prediction from the graph branch while incorporating holistic image context from the global branch. It helps maintain reliable grading when lesions are sparse, missed, or unreliably segmented.

\subsection{Graph Reasoning and Optimization}
\label{sec:reason}
The assembled graph is reasoned over by a GAT~\cite{gat}. Each node $n_j \in \mathcal{V}$ is assigned a confidence score $s_j \in [0,1]$ indicating its reliability, and edge-wise gating weights are computed from the confidence of both connected nodes to control information propagation.

For \textit{lesion nodes}, the confidence is defined as the prediction score from the lesion segmentation model. 
For \textit{intermediate nodes}, it is computed as the average confidence of their associated lesion nodes, reflecting aggregated reliability at categorical semantic or regional level.
For the \textit{anatomical node}, confidence is aggregated from its associated categorical nodes, propagating lesion-level reliability up to the anatomical level.
As for the \textit{image-level semantic node}, the confidence score is set to $1$, as it is independent of lesion segmentation. 

The message passed from node $j$ to node $k$ is gated by both endpoints,
\begin{equation}
\label{eq:gate}
m_{jk} = g(s_j)\cdot g(s_k)\cdot \alpha_{jk}\cdot \mathbf{W}(h_j),
\qquad g(s)=\sigma(\lambda s+\beta),
\end{equation}
where $\alpha_{jk}$ is the softmax-normalized GAT attention weight~\cite{gat}, $\mathbf{W}$ a learnable projection, $h_j$ the feature of node $j$, and $\lambda,\beta$ learnable gating parameters.

Each node is then updated by aggregating its gated messages, $h_k' = \sigma\big(\sum_{j\in\mathcal{N}(k)} m_{jk}\big)$. After $L$ GAT layers, five learnable query tokens attend to high-level nodes (anatomical, semantic, categorical, and regional) via multi-head cross-attention. The resulting features are concatenated and projected to obtain the graph-branch logits $\ell^{g}$, which are fused with the global branch via Eq.~\ref{eq:fusion}. The framework is trained end-to-end with a cross-entropy objective on $\hat{p}$.

\section{Experiments}

\begin{table*}[t]
\centering
\caption{Performance comparison on Messidor-2, APTOS, and DDR (all metrics $\times 100$). Best results are shown in \textbf{bold}, and second-best results are underlined.}
\label{tab:main_results}

\footnotesize
\setlength{\tabcolsep}{3pt}
\renewcommand{\arraystretch}{1.15}

\begin{tabular}{llccccccccc}
\toprule

Dataset & Metric
& PARAGraph & QAGCN & GREEN
& Swin-L & ViT
& EfficientNet & DenseNet & CABNet
& RetFound \\

\midrule

\multirow{4}{*}{Messidor}

& QWK
& \textbf{88.47$\pm$1.03} & 41.49$\pm$11.91 & 36.24$\pm$4.32 & \underline{87.82$\pm$1.87} & 83.23$\pm$1.28 & 85.29$\pm$0.94 & 82.88$\pm$2.12 & 84.19$\pm$1.90 & 71.16$\pm$3.72 \\

& ACC
& \textbf{80.46$\pm$1.25} & 56.26$\pm$3.75 & 45.23$\pm$4.30 & \underline{78.16$\pm$1.63} & 73.28$\pm$2.31 & 75.46$\pm$2.17 & 73.85$\pm$1.59 & 75.52$\pm$3.45 & 59.94$\pm$4.48 \\

& F1
& \underline{69.60$\pm$3.08} & 29.35$\pm$6.17 & 30.48$\pm$5.09 & \textbf{71.96$\pm$2.19} & 65.45$\pm$2.63 & 65.40$\pm$2.98 & 66.27$\pm$3.52 & 67.83$\pm$3.88 & 54.72$\pm$3.19 \\

& SPE
& \textbf{93.62$\pm$0.45} & 86.19$\pm$1.67 & 85.42$\pm$0.79 & \underline{93.46$\pm$0.35} & 91.33$\pm$0.64 & 92.19$\pm$0.40 & 91.67$\pm$0.48 & 92.37$\pm$0.63 & 88.45$\pm$0.77 \\

\midrule

\multirow{4}{*}{APTOS}

& QWK
& \textbf{93.85$\pm$0.25} & 90.84$\pm$0.73 & 78.47$\pm$2.91 & 91.16$\pm$1.19 & 91.18$\pm$1.49 & 89.03$\pm$1.38 & 90.67$\pm$1.48 & 90.59$\pm$1.32 & \underline{91.53$\pm$1.54} \\

& ACC
& \textbf{88.61$\pm$0.55} & 84.03$\pm$1.71 & 75.65$\pm$0.85 & \underline{87.50$\pm$0.85} & 86.00$\pm$2.20 & 84.65$\pm$2.15 & 86.05$\pm$1.84 & 87.25$\pm$0.89 & 86.50$\pm$2.10 \\

& F1
& \textbf{75.43$\pm$1.05} & 65.44$\pm$3.81 & 49.80$\pm$0.79 & \underline{68.80$\pm$2.53} & 65.90$\pm$3.48 & 60.58$\pm$3.93 & 66.36$\pm$4.49 & 66.99$\pm$1.87 & 65.56$\pm$5.41 \\

& SPE
& \underline{97.06$\pm$0.12} & 96.11$\pm$0.37 & 93.77$\pm$0.36 & \textbf{97.13$\pm$0.18} & 96.62$\pm$0.49 & 96.24$\pm$0.60 & 96.81$\pm$0.37 & 96.92$\pm$0.27 & 96.78$\pm$0.54 \\

\midrule

\multirow{4}{*}{DDR}

& QWK
& \textbf{88.62$\pm$0.20} & 72.29$\pm$3.07 & 69.72$\pm$0.66 & \underline{87.36$\pm$0.84} & 81.08$\pm$0.76 & 83.94$\pm$1.11 & 79.97$\pm$1.01 & 82.89$\pm$0.73 & 63.75$\pm$1.96 \\

& ACC
& \textbf{84.99$\pm$0.50} & 65.75$\pm$2.27 & 61.42$\pm$2.55 & \underline{84.30$\pm$0.86} & 71.87$\pm$4.19 & 82.31$\pm$0.87 & 78.93$\pm$1.57 & 81.45$\pm$0.37 & 65.39$\pm$5.51 \\

& F1
& \underline{71.10$\pm$1.12} & 48.55$\pm$3.07 & 47.71$\pm$1.33 & \textbf{72.58$\pm$1.14} & 58.34$\pm$2.39 & 66.02$\pm$1.53 & 65.44$\pm$1.63 & 65.34$\pm$1.17 & 43.41$\pm$4.32 \\

& SPE
& \textbf{95.43$\pm$0.15} & 89.33$\pm$0.78 & 88.72$\pm$0.30 & \underline{95.16$\pm$0.29} & 92.14$\pm$0.57 & 94.39$\pm$0.31 & 93.32$\pm$0.34 & 93.98$\pm$0.23 & 89.11$\pm$0.61 \\

\bottomrule
\end{tabular}
\end{table*}

\begin{table}[t]
\centering
\caption{Ablation on DDR (QWK $\times 100$, mean over five runs). Top: progressive graph construction; bottom: component removal. $\Delta$ is relative to the OD+global baseline (top) or the full model (bottom). The flat and full graph variants have identical parameter counts (95.1M).}
\label{tab:ablation}
\small
\begin{tabular}{lcc}
\toprule
Variant & QWK & $\Delta$ \\
\midrule
OD + global (no lesion graph)          & 87.85 & -- \\
\quad + lesion nodes (flat graph)      & 88.00 & +0.15 \\
\quad + categorical nodes               & 88.31 & +0.46 \\
\quad + regional nodes (Full)           & \textbf{88.62} & +0.77 \\
\midrule
Full                                    & 88.62 & -- \\
\quad $-$ OD positional encoding        & 88.29 & $-0.33$ \\
\quad $-$ medical-prior fusion          & 87.74 & $-0.88$ \\
\quad $-$ late fusion (global logits)   & 88.23 & $-0.39$ \\
\quad $-$ global pathway (early+late)   & 86.80 & $-1.82$ \\
\bottomrule
\end{tabular}
\end{table}

\subsection{Experimental Setup}

\subsubsection{Datasets}

We evaluate PARAGraph on three public DR grading datasets: DDR~\cite{ddr}, Messidor-2~\cite{messidor1,messidor2}, and APTOS~\cite{aptos}. For DDR, we retain the official test set and merge the official training and validation sets after removing ungradable images. The merged data are then stratified into 80\% training and 20\% validation subsets using five random seeds (42, 0, 1, 2, and 1024). Messidor-2 and APTOS contain 1{,}744 and 2{,}013 gradable images, respectively, after excluding low-quality APTOS images. Without official splits, both datasets are randomly divided into training and test sets at a 4:1 ratio, without separate validation sets. We report the mean performance over five runs for all datasets.

\subsubsection{Evaluation Metrics and Baseline Methods}

We report QWK, ACC, macro-F1, and SPE, measuring ordinal agreement, overall accuracy, class-balanced performance, and specificity, respectively. Eight baselines are considered: DenseNet-121~\cite{densenet}, EfficientNet-B7~\cite{efficientnet}, CABNet~\cite{cabnet}, ViT-Large~\cite{vit}, Swin-Large~\cite{swin}, RetFound~\cite{retfound}, GREEN~\cite{green}, and QAGCN~\cite{qagcn}. Unlike the class-dependency GREEN and image-level QAGCN, PARAGraph models intra-retinal anatomical relationships. We did not identify publicly available official implementations of the closely related PRANet and MDGNet, so we do not report unverified reproductions of those methods. Most baselines use $1024\times1024$ inputs, except RetFound and QAGCN at $224\times224$. PARAGraph uses $1024\times1024$ for segmentation and $512\times512$ for Swin feature extraction. The relatively low performance of GREEN and QAGCN on Messidor-2 is associated with slow convergence and training instability, respectively, under the limited-data setting.

\subsubsection{Implementation Details}

Images are bilinearly resized to $1024\times1024$. All experiments use PyTorch 2.8.0 on a single NVIDIA RTX 5090. PARAGraph is trained using AdamW with batch size 8, initial learning rate $10^{-4}$, and cosine annealing. Lesion, OD, and fovea segmentation models, pretrained on IDRiD~\cite{idrid} and FGADR~\cite{fgadr}, remain frozen. Lesion nodes are localized from UNet~\cite{unet} masks, with embeddings extracted by grid-sampling multi-scale OD-anchored encoder features at lesion centers, followed by per-scale projection and fusion. The anatomical node is similarly encoded at the OD center.

\subsection{Experimental Results}

\begin{figure}[!t]
  \centering
  \includegraphics[width=\linewidth]{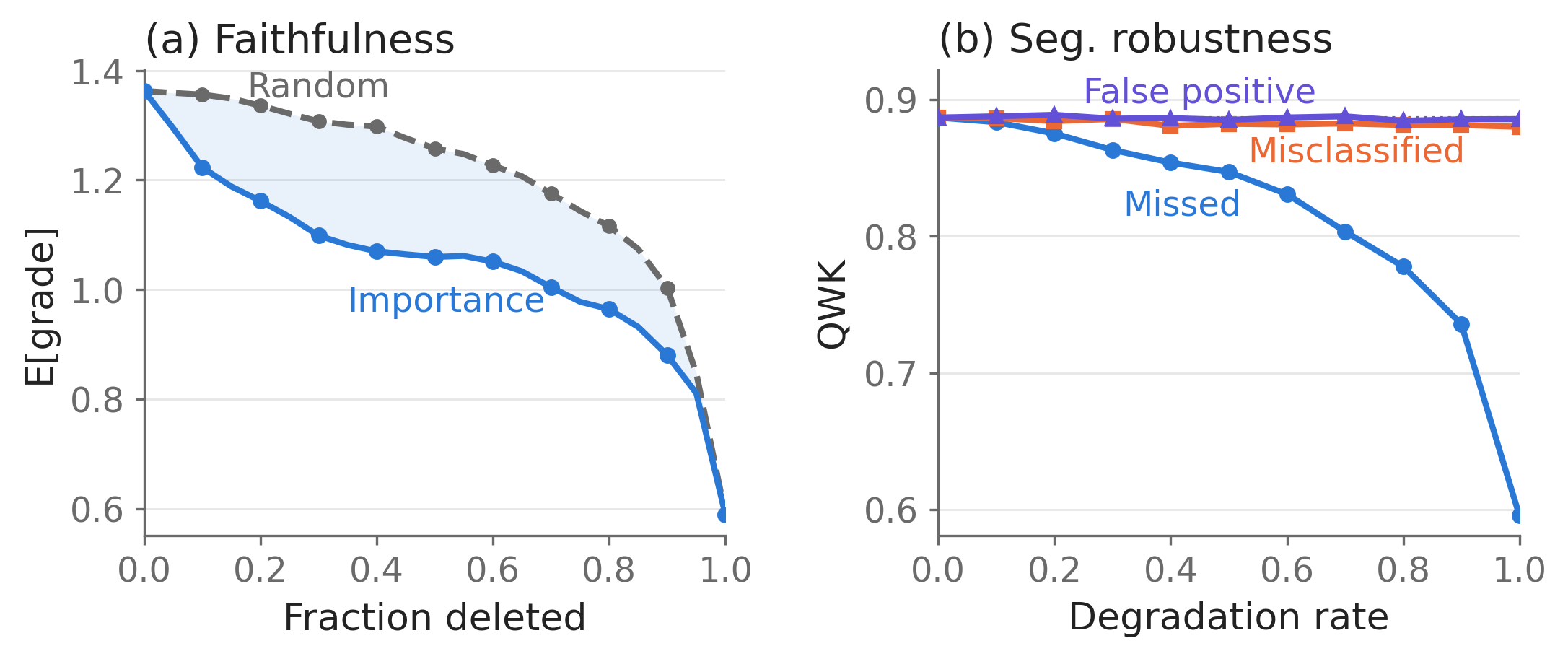}
  \caption{Faithfulness and segmentation robustness on DDR. (a) Deleting highly ranked lesions reduces the expected grade faster than random deletion. (b) QWK under increasing missed-lesion, lesion-category, and false-positive perturbations.}
  \label{fig:credibility}
\end{figure}

\begin{figure}[!t]
  \centering
  \includegraphics[width=\linewidth]{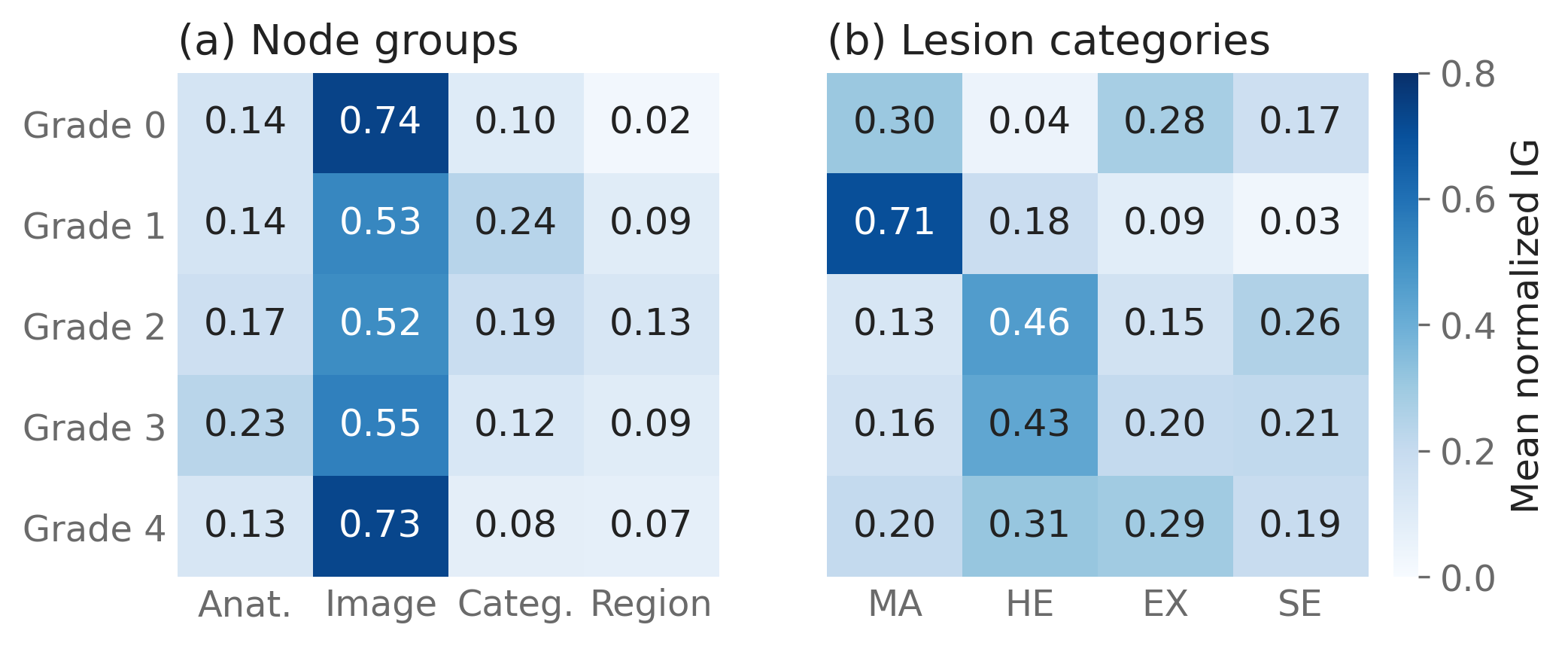}
  \caption{Integrated-Gradients node attribution per grade (correctly classified test cases). \emph{Left}: high-level node groups; \emph{right}: lesion-category attribution. Microaneurysms dominate Grade~1, while hemorrhages dominate Grades~2--3 among lesion categories, consistent with clinical grading criteria. Global semantic features provide complementary context, particularly for Grades~0 and~4.}
  \label{fig:ig}
\end{figure}

\begin{figure*}[!t]
    \centering
    \includegraphics[width=0.75\textwidth]{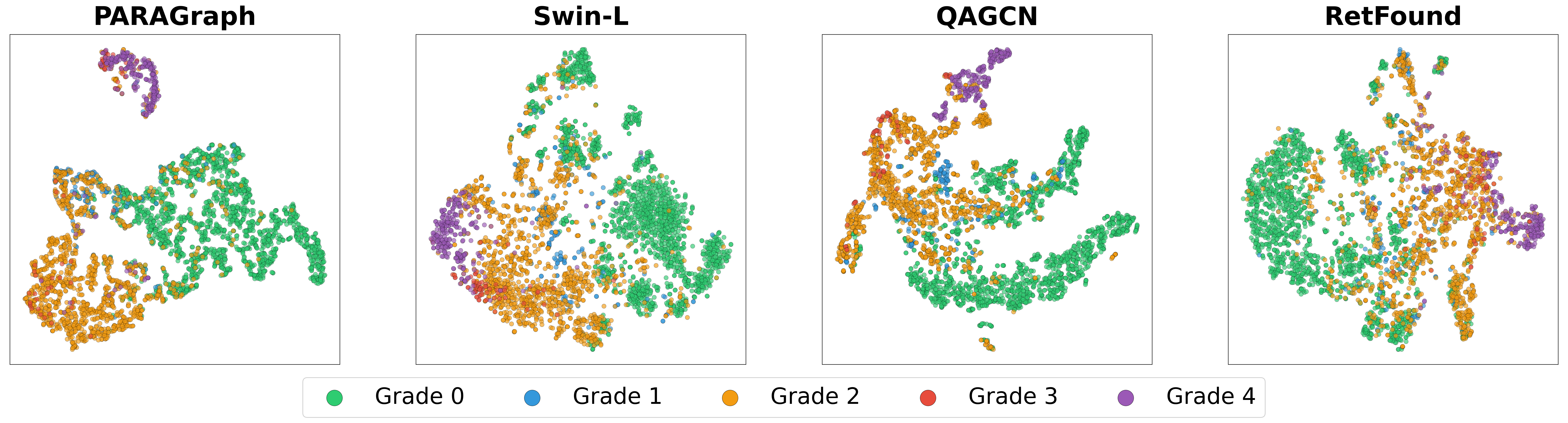}
    \caption{t-SNE visualization of feature embeddings on the DDR dataset.}
    \label{fig:tsne_comparison}
\end{figure*}

\subsubsection{Quantitative Results}

Table~\ref{tab:main_results} reports results on Messidor-2, APTOS, and DDR. PARAGraph achieves the best QWK on all three datasets, with scores of $88.47$, $93.85$, and $88.62$, outperforming the second-best method by $+0.65$, $+2.32$, and $+1.26$, respectively. The gains are especially clear on APTOS and DDR, suggesting better preservation of the ordinal severity structure.

PARAGraph also achieves the highest ACC on all datasets and the highest SPE on Messidor-2 and DDR. For macro-F1, it ranks first on APTOS by a clear margin and second on Messidor-2 and DDR. Compared with the graph-based baselines GREEN and QAGCN, PARAGraph is consistently stronger across datasets, highlighting the potential of explicit intra-retinal graph reasoning over class-level or image-level graph modeling.

PARAGraph further substantially outperforms RetFound on DDR, with $88.62$ versus $63.75$ QWK. Under our evaluation protocol, PARAGraph reasons over high-resolution lesion evidence within an anatomy-aware graph, although differences in input resolution and training settings should be considered when interpreting cross-model gaps. These consistent improvements validate the effectiveness of the proposed framework.

\subsubsection{Ablation Study}
We ablate each component of PARAGraph on DDR, the largest benchmark, and report the mean QWK over five runs with different training/validation splits (Table~\ref{tab:ablation}).

\textbf{Hierarchical graph and clinical priors.}
Starting from the OD+global baseline (87.85 QWK), connecting lesion nodes directly to both hubs forms a flat graph and increases QWK to 88.00. Adding categorical and regional aggregation raises QWK by a further 0.31 points each, reaching 88.62. Thus, hierarchical aggregation improves upon the flat graph by 0.62 points with the same overall parameter count (95.1M), supporting the value of structured lesion reasoning beyond direct connections. Removing medical-prior fusion reduces QWK by 0.88 points, demonstrating the importance of anatomy-aware lesion representations.

\textbf{Global context fusion.}
Removing late fusion, i.e., the global-branch logits, reduces QWK by $0.39$, while removing the entire global pathway reduces it by $1.82$. This suggests that early fusion contributes more substantially, and that early and late fusion are complementary. Together, they provide holistic context beyond lesion-grounded evidence.

\textbf{OD-anchored positional encoding.}
Disabling the OD-anchored positional bias lowers QWK across all five splits, with an average drop of $0.33$, confirming that anatomy-anchored encoding provides useful spatial information for node feature extraction.

\subsubsection{Interpretability and Robustness}

\textbf{Lesion-level faithfulness.}
We evaluate lesion-level faithfulness on 56 DDR test images with pixel-level ground-truth masks. Replacing predicted masks with ground truth, we measure each lesion's leave-one-out importance by its effect on the expected grade and analyze lesion-deletion curves~\cite{rise} and top-lesion retention (Fig.~\ref{fig:credibility}). Removing highly ranked lesions degrades predictions faster than random deletion (deletion-gap AUC: $0.100$). Removing the top-$20\%$ lesions reduces predicted-class probability by $0.080$, whereas retaining only these lesions changes it by just $0.015$~\cite{eraser}. These results support the faithfulness of the identified lesion evidence.

\textbf{Robustness to segmentation degradation.}
We inject controlled segmentation errors to evaluate robustness (Fig.~\ref{fig:credibility}). As lesion miss rates increase from $0$ to $100\%$, QWK decreases from $88.7$ to $59.5$, with substantial degradation under severe lesion omissions. In contrast, lesion-category errors and false positives have limited impact, with QWK remaining near $88$. This indicates robustness to these perturbations but sensitivity to extensive missed lesions.

\textbf{Node attribution.}
Integrated Gradients~\cite{ig} reveals clinically aligned lesion-category attribution patterns (Fig.~\ref{fig:ig}). Microaneurysms dominate Grade~1, consistent with mild NPDR, while hemorrhages become more influential in Grades~2--3, aligning with more advanced NPDR. These patterns emerge without explicitly imposing grade-specific lesion weights, suggesting that PARAGraph naturally learns clinically meaningful lesion importance. Meanwhile, global semantic nodes provide complementary context, particularly for Grades~0 and~4, where segmented lesion evidence may be limited.

\textbf{Feature Space Visualization.}
t-SNE visualization of DDR test embeddings (Fig.~\ref{fig:tsne_comparison}) shows more compact and structured grade clusters for PARAGraph than for Swin-L, QAGCN, and RetFound. Grade 0 is clearly separated, Grades 1--3 form a progressive NPDR continuum, and Grade 4 forms a distinct cluster. This qualitative separation suggests that PARAGraph learns more structured grade representations, complementing the quantitative and lesion-attribution analyses.

\section{Conclusion}
In this paper, we presented \textbf{PARAGraph}, a pathology-anatomy-aware hierarchical graph for DR grading that integrates clinical priors, OD-anchored positional encoding, and global context. Experiments on Messidor-2, APTOS, and DDR demonstrate superior performance over CNN, transformer, graph-based, and foundation-model baselines. Faithfulness and robustness analyses confirm clinically meaningful lesion attribution and resilience to segmentation errors. Lesion-category attribution further reveals clinically coherent patterns, with microaneurysms most influential in mild NPDR and hemorrhages in moderate-to-severe NPDR. Future work will explicitly model neovascularization and investigate cross-domain generalization. Overall, PARAGraph demonstrates the potential of clinically grounded graph reasoning for trustworthy DR grading.

\section*{Acknowledgment}
This work was supported in part by the St Paul's Eye Research Foundation (Grant No. CF349 25.26). The authors acknowledge the LaTIM laboratory for making the Messidor-2 database available. The Messidor-Original subset was kindly provided by the Messidor program partners (see \url{https://www.adcis.net/en/third-party/messidor/}).

\bibliographystyle{IEEEtran}
\bibliography{references}

\end{document}